\documentclass[10pt,journal,compsoc]{IEEEtran}
\usepackage[dvipsnames]{xcolor}
\usepackage{multicol, blindtext}

\usepackage[style=ieee,maxcitenames=1,mincitenames=1]{biblatex}
\ifCLASSINFOpdf
  \usepackage[pdftex]{graphicx}
  \graphicspath{{./images/}}
\else
\fi
\usepackage{amsmath}
\usepackage{mathrsfs}
\usepackage{bbold}

\usepackage{tabu}
\usepackage{multirow}
\usepackage{booktabs}

\usepackage{array}
\ifCLASSOPTIONcompsoc
 \usepackage[caption=false,font=footnotesize,labelfont=sf,textfont=sf]{subfig}
\else
 \usepackage[caption=false,font=footnotesize]{subfig}
\fi
\usepackage{caption}
\usepackage{fixltx2e}
\begin{document}
\title{Learning Global and Local Consistent Representations for Unsupervised Image Retrieval via Deep Graph Diffusion Networks}%

\author{
Zhiyong Dou, %
        Haotian Cui, %
        Lin Zhang,
        and~Bo Wang,~\IEEEmembership{Member,~IEEE,}%

\IEEEcompsocitemizethanks{\IEEEcompsocthanksitem Z. Dou is with the University of Toronto, Toronto, Ontario M5S 1A1, Canada. He is also with the School of Electronic Information and Communications, Huazhong University of Science and Technology, Wuhan, Hubei 430074, China.\protect\\
E-mail: zydou@hust.edu.cn

\IEEEcompsocthanksitem H. Cui is with the University of Toronto, Toronto, ON M5S 1A1, Canada. He is also with Vector Institute, Toronto, ON M5G 1M1, Canada. \protect\\
E-mail: htcui@cs.toronto.edu

\IEEEcompsocthanksitem L. Zhang is with the University of Toronto, Toronto, ON M5S 1A1, Canada.\protect\\
E-mail: linzhang@utstat.toronto.edu

\IEEEcompsocthanksitem B. Wang is with Peter Munk Cardiac Centre, Toronto, ON M5G 2N2, Canada. He is also with Vector Institute, Toronto, ON M5G 1M1 and the University of Toronto, Toronto, ON M5S 1A1, Canada. \protect\\
E-mail: bo.wang@uhnresearch.ca
}%
\thanks{Zhiyong Dou and Haotian Cui contribute equally. Corresponding author: Bo Wang. E-mail: bo.wang@uhnresearch.ca}
}

\IEEEtitleabstractindextext{%
\begin{abstract}
Diffusion has shown great success in improving the accuracy of unsupervised image retrieval systems by utilizing high-order structures of image manifold. However, existing diffusion methods suffer from three major limitations: 1) they usually rely on local structures without considering global manifold information; 2) they focus on improving pairwise similarities within existing input images
\textbf{\emph{transductively}} while lacking the flexibility to learn representations for unseen instances \textbf{\emph{inductively}}; 3) they fail to scale to large datasets due to prohibitive memory consumption and computational burden due to the intrinsic high-order operations on the whole graph. In this paper, to address these limitations, we propose a novel method, \textbf{\emph{Gra}}ph \textbf{\emph{D}}iffusion \textbf{\emph{Net}}works (GRAD-Net), that adopts graph neural networks (GNNs), a novel variant of deep learning algorithms on irregular graphs. GRAD-Net learns semantic representations by exploiting both local and global structures of image manifold in an unsupervised fashion. By utilizing sparse coding techniques, GRAD-Net not only preserves global information on the image manifold, but also enables scalable training and efficient querying.
Experiments on several large benchmark datasets demonstrate the effectiveness of our method over existing state-of-the-art diffusion algorithms for unsupervised image retrieval. 
\end{abstract}

}

\maketitle

\IEEEdisplaynontitleabstractindextext
\IEEEpeerreviewmaketitle

\IEEEraisesectionheading{\section{Introduction}\label{sec:introduction}}

\IEEEPARstart{U}{nsupervised} image retrieval refers to the task of returning relevant instances in a database given an unlabeled query, which is an important basis for various applications such as information search and database management \cite{ghorab2013personalised,van2017automatic}. Traditionally, unsupervised image retrieval is accomplished by computing either predefined pairwise similarities (e.g. Euclidean distances) or by adopting a random-walk style process (e.g. diffusion \cite{donoser2013diffusion}). It is widely acknowledged that no single metric can generate reliable retrieval performance due to the ``curse'' of dimensionality \cite{bach2017breaking}. Consequently, most of the existing works on unsupervised image retrieval have been focusing on exploiting the diffusion process to learn context-sensitive affinity measures \cite{iscen2017efficient,bai2018regularized,wang2013dynamic}. 

\begin{figure}[h]
\centering
\includegraphics[width=0.4\textwidth]{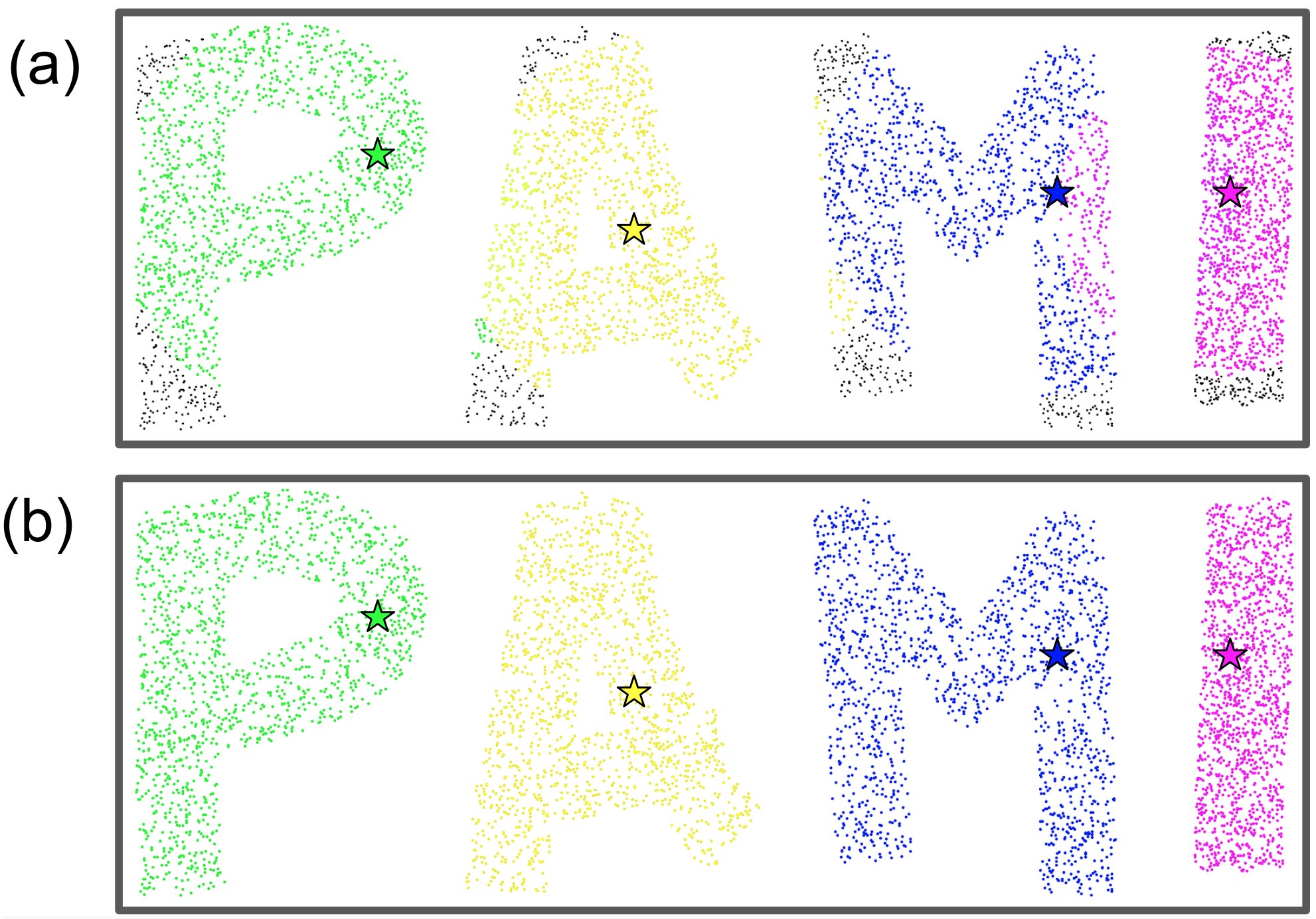}
\caption{{\bf The retrieval results on a synthetic toy dataset} using (a) Euclidean distance and (b) diffusion. Four queries are marked as stars.}
\label{fig:toy}
\end{figure}

The principle of capturing geometric manifold applies to most existing diffusion methods. First, the manifold is interpreted as a weighted graph, where each instance is represented by a node, and weights on the edges connecting nodes represent the similarities. The pairwise affinities are then updated iteratively by diffusing along the graph geometry. This diffusion process, originally proposed by \textcite{zhou2004ranking}, typically follows the concept of random walk, where a transition matrix derived from the edge weights determines the probabilities of transiting from one node to another. The updated affinity values in turn improve the retrieval results. To illustrate this concept, we consider the following toy example. As shown in Fig. \ref{fig:toy}, each point is a simple analogy to the feature vector of an image. The data points consist of four letters, each with 1,500 points. The queries are displayed as the star nodes. The ideal result for this retrieval is that the points from the same letter as the query point should have the highest ranks. Euclidean distance as a similarity metric is inadequate for this task (Fig. \ref{fig:toy}a), while in comparison, after diffusing the similarities on the graph, the retrieval result is significantly improved (Fig. \ref{fig:toy}b).

The success of deep neural networks on various applications partly results from the powerful image features with rich semantic information. Models, particularly deep convolutional networks \cite{babenko2014neural,gordo2016deep,radenovic2016cnn, razavian2016visual} pre-trained on large datasets,  such as ImageNet \cite{deng2009imagenet} and Landmarks \cite{babenko2014neural}, are increasingly being used for feature extraction. Diffusion is then deployed on the extracted features to further improve the retrieval results.

However, these diffusion-based methods suffer from three main limitations. First, they usually rely on local structures without considering global manifold information, that is sparsified graphs based on neighborhood search are extensively used without considering the \textit{cluster} structures as a whole. Secondly, these diffusion-based methods focus on improving pairwise similarities within existing images \emph{transductively} but lack the flexibility to learn representations for unseen instances \emph{inductively}. In other words, it is difficult to generalize the diffusion models to unobserved databases without re-computing the diffusion iterations. Lastly, they usually fail to scale to large datasets due to prohibitive memory consumption and computational burden resulting from the intrinsic high-order operations on the whole graphs.

Graph neural networks (GNNs) have recently emerged as a promising line of research that adopts convolutional operators on irregular inputs like graphs \cite{kipf2016semi,atwood2016diffusion,niepert2016learning,cao2016deep}. For instance, there has been an increasing interest in applying GNNs to citation network analysis \cite{kipf2016semi,levie2018cayleynets}, collaborative filtering \cite{wang2019neural} and knowledge graphs \cite{park2019estimating}, all of which demonstrated state-of-the-art performance. However, most existing GNNs rely on supervised learning to train an effective model, which prevents direct adoption of GNNs to the unsupervised image retrieval tasks due to the lack of labeled data. 

In this paper, we propose a novel approach, \textbf{\emph{Gra}}ph \textbf{\emph{D}}iffusion \textbf{\emph{Net}}works (GRAD-Net), for unsupervised image retrieval. GRAD-Net consists of GNNs that are trained with two loss functions that directly depict the diffusion process without any labeled information. The diffusion process is thus fully learnable, and the features are trained to transform on an optimal manifold subject to the defined loss in an unsupervised fashion. Benefiting from the generalizability of GNNs, our model can be easily extended to unseen queries. Furthermore, to consider the global structures, GRAD-Net constructs an additional bipartite graph to learn consistent global and local representations of all the images. GRAD-Net alleviates the common scalability issue of diffusion methods by effectively sampling sub-networks in a mini-batch manner at each training iteration. We demonstrate the effectiveness of GRAD-Net with extensive empirical experiments.

Our main contributions are summarized as follows:
\begin{enumerate}
  \item \textbf{Unsupervised representation learning on image manifolds}. GRAD-Net, among the first-in-class deep learning models for unsupervised diffusion process on image retrieval, updates the instance features (i.e. the node attributes on the graphs) in an end-to-end manner rather than the conventional choice of diffusion rank values or similarities on manifolds. We transform the conventional diffusion from a message-passing process to an optimization process through a multi-layer neural networks with an efficient diffusion-like operator. We show that this approach enriches the features with more semantic information, and the learned features can be used to perform efficient retrieval with a simple nearest neighbor search and support effective inductive learning. 
  \item \textbf{Achieving global and local consistency}. We introduce a series of novel techniques including: (i) second order propagation that accelerates the message aggregation in GNNs; (ii) local and global loss functions that leverage the structure of the manifold; (iii) sparse coding that adopts a bipartite graph technique on top of the sparsified pairwise graphs to reflect the global \emph{cluster} information. These novel techniques collectively contribute to GRAD-Net to achieve global and local consistency.
  \item \textbf{Intrinsic scalability and high efficiency}. Unlike traditional diffusion methods that suffer from at least quadratic computational complexity, our proposed GRAD-Net has linear complexity and is intrinsically scalable to large databases through mini-batch training on the graphs. GRAD-Net has proven its high efficiency in both training and query processes on very large-scale datasets. 
\end{enumerate}

\section{Related Work}
\subsection{Diffusion for Unsupervised Image Retrieval}

Metric learning has been a central task for image retrieval. There is a myriad of works focusing on learning a general \emph{metric} to compute pairwise distances between images \cite{BMVC2016_119}, \cite{yu2018hard}. In contrast to conventional metric learning methods, diffusion-based methods have emerged as a promising alternative that learns pairwise similarities on image manifolds. Introduced by \textcite{zhou2004ranking}, diffusion has shown consistent improvements over raw distances/similarities by exploiting intrinsic manifold geometry \cite{zhou2004learning}. Inspired by semi-supervised learning, Graph Transduction (GT) \cite{bai2009learning} takes the query point as the only labeled data, and propagates the labeled point to unlabeled databases in a similar fashion to label propagation \cite{raghavan2007near}. Motivated by the observation that a good ranking is usually asymmetrical, Contextual Dissimilarity Measure (CDM) \cite{jegou2008accurate} improves Bag-of-Words (BoW) \cite{fei2005bayesian} retrieval system by iteratively estimating the pairwise distance in the spirit of Sinkhorn’s scaling algorithm \cite{sinkhorn1964relationship}. 

Further, noticing that diffusion is susceptible to noise edges in the affinity graph, Locally Constrained Diffusion Process (LCDP) \cite{wang2012affinity} stresses that it is crucial to constrain the diffusion process ``locally". Along the same line, Tensor Product Graph diffusion (TPG) \cite{yang2012affinity} manages to leverage the high-order information from the tensor product of the affinity graphs. However, TPG constrains the pathways of message passing to local neighbors only such that the computational complexity does not increase significantly. A detailed comparison of various diffusion methods has been conducted in a survey \cite{donoser2013diffusion} by enumerating 72 variants of diffusion process (4 different affinity initializations, 6 different transition matrices and 3 different updating schemes). \textcite{bai2017regularized} provides empirical and theoretical results and suggests that metric learning on tensor product graphs with high-order information is more robust to retrieval. Recently, \textcite{bai2018regularized} enhances the tensor product to a regularization process on graphs and displays its potential of retrieval for heterogeneous instances.

\subsection{Graph Neural Networks}
Recent years have witnessed an increasing trend in applying deep learning algorithms to irregular inputs such as graphs. Early works include recursive neural networks \cite{frasconi1998general} that represent and process graph data. GNNs were introduced by \textcite{gori2005new,scarselli2008graph} as a generalization of recursive neural networks that can directly deal with graphs. Typical GNNs consist of an iterative random-walk process in which node states are propagated based on certain probability distribution until convergence. This idea was improved by \textcite{li2015gated} that uses gated recurrent units \cite{cho2014learning} in the propagation step. However, these recurrent models usually suffer from large computational burdens and therefore hard to train and scale to large graphs. 

Inspired by the tremendous success of convolutional neural networks (CNNs) in computer vision, \textcite{bruna2013spectral} introduced the convolution operation in the Fourier domain by computing the eigendecomposition of the graph Laplacian. It was then improved by \textcite{henaff2015deep} that parameterizes the spectral filters with smooth coefficients such that the computations are spatially localized. The milestone work by \textcite{kipf2016semi} introduced graph convolutional network (GCN) that simplifies the previous methods by restricting the filters to operate in the first-order neighborhood around each node. Since then, various modifications have been proposed to improve upon GCN. For example, \textcite{velivckovic2017graph} adopted the attention mechanism to address the high sensitivity to noise issue in the graph edges of GCN by computing learnable attentions among the nodes.

The applications of GNNs in computer vision are booming with notable examples such as few-shot image classification \cite{garcia2017few, guo2018neural}, semantic segmentation \cite{qi20173d, yi2017syncspeccnn} and visual question answering \cite{chen2018iterative, narasimhan2018out}. However, most of these applications require an extensive amount of labeled data. Recently, \textcite{jiang2019data} introduces new GNN layer operations for feature diffusion in a semi-supervised setting, and Guided Similarity Separation (GSS) \cite{liu2019guided} applies a Kipf-style GCN \cite{kipf2016semi} to unsupervised image retrieval but lacks the scalability to large datasets. Apart from the aforementioned works, few has focused on unsupervised image retrieval.

The proposed GRAD-Net, to our best knowledge, is one of the first attempts to adopt GCN-like algorithms in unsupervised image retrieval. Our proposed network requires no label information yet extends the traditional diffusion models to learn global and local consistent representations and enables effective training and querying on large-scale datasets. 

\section{Our Approach}
In this section, we introduce the detailed components of the proposed Graph Diffusion Network (GRAD-Net). We first review the theoretical bases of the diffusion process on graphs and GCN, and then elaborate on the structures of our proposed GRAD-Net. Though our proposed method can be easily generalized to the retrieval of various types of data, in this work, we will focus on image retrieval to illustrate the idea of GRAD-Net.

\subsection{Problem Setup and Notations} \label{sec:setup}
For image retrieval tasks, we denote the dataset as $\mathcal{X} = \{x_1,x_2,\ldots,x_n\}\subset R^d$, where $x_i$ is a feature vector of an image. We assume the set of queries can be projected to the same feature space as the instances in the dataset, and denote the queries as $Q=\{q_1,q_2, \ldots,q_m\}\subset R^d$, where $q_i$ represents the feature vector of the $i$-th query image and $m$ is the number of query images. For simplicity, but without loss of generality, we consider the scenario that the instance (or query) is interpreted as a single vector. The combined set of dataset instances and queries can then be expressed as 

\begin{equation} \label{eq:input}
    \bar{\mathcal{X}} = \mathcal{X} \cup Q = \{q_1,...,q_m,x_1,...,x_n\},
\end{equation}
and we use $\bar{\mathcal{X}}_i$ denotes the \textit{i}-th instance in $\bar{\mathcal{X}}$.

\subsection{Diffusion on Graphs} \label{sec:diff}
A graph $G=\{V, E\}$ consists of a set of nodes $V$ and edges $E$. An affinity matrix $A$ represents the adjacent node pairs in the graph. Two main approaches to conduct diffusion on graphs are (i) iteratively updating the pairwise similarities \cite{zhou2004ranking, donoser2013diffusion} and (ii) solving the closed form of Eq.\ref{eq:f*} directly \cite{iscen2017efficient}. However, both approaches are essentially based on the same random walk mechanism proposed by \textcite{zhou2004ranking}.

To perform a random walk on a graph $G$, a transition matrix is introduced to describe the probability of walking from one node to another, which is considered to be proportional to the affinity value. A degree matrix $D$ is introduced to normalize the affinity matrix, which produces the transition matrix. The degree matrix 
$$
D_{i,j}:= \begin{cases}
			\sum_{k=1}^{n+m}a_{ik}, & \text{if }i=j\\
            0, & \text{otherwise}
		 \end{cases}
$$
is a diagonal matrix with each diagonal element corresponding to the row-wise sum of a predefined affinity matrix $A$. Then the transition matrix $S$ is computed as:
\begin{equation} \label{eq:s}
    S=D^{- 1/2}AD^{- 1/2}.
\end{equation}

Based on the transition matrix $S$, random walk is then performed on the graph to update a state vector $f^t \in R^{n+m}$ until it converges. The process iterates at the \textit{t}-th step of the random walk as follows, 
\begin{equation} \label{eq:rw}
    f^{t+1} = \alpha Sf^t + (1-\alpha)f^0, \alpha \in (0,1),
\end{equation}
where $f^t = [{f^t_q}^T,{f^t_d}^T]^T$ is composed of both the query state $f^t_q\in R^m$ and the dataset state $f^t_d\in R^n$. The initial state $f^0$ is a binary vector set to $f^0 = [{f^0_q}^T, {f^0_d}^T]^T =[\textbf{1}^T, \textbf{0}^T]^T$. Eq.\ref{eq:rw} can be intuitively interpreted as follows. Given a state $f^t$, it transits based on $S$ with a probability of $\alpha$ and restarts from the initial state with a probability of $(1-\alpha)$. This process will converge to a closed-form solution \cite{zhou2004ranking}:
\begin{equation} \label{eq:f*}
    f^* = (1-\alpha)(I-\alpha S)^{-1}f^0
\end{equation}

The final state, $f^{t=T}$, obtained after certain iterations or derived directly from the closed-form solution, represents the similarities between the dataset instances and the queries, which eventually determines the rank of the dataset instances.

However, Eq.\ref{eq:rw} does not explicitly consider the high-order information on the manifold. \textcite{yang2012affinity} addressed this issue and introduced a modified diffusion on a tensor product graph where nodes refer to the instance pairs, which naturally takes the higher-order information into account. The iteration step in the Tensor Product Graph diffusion (TPG diffusion) is:
\begin{equation} \label{eq:tp}
    \hat{A}^{t+1} = S\hat{A}^tS^T+I.
\end{equation}
The final $\hat{A}^{t=T} \in R^{n\times n}$ after iterations stands for the updated affinity matrix. As argued by \textcite{donoser2013diffusion}, the TPG diffusion achieved the most robust performance among all the compared methods. However, TPG diffusion suffers from heavy computational burden when handling large networks due to the large-scale tensor product graph. Our proposed method adopts a second-order operator similar to TPG diffusion, but overcomes the scalability issue by applying the second-order operator on a mutual k-NN graph (see detailed description in Section \ref{sec:gdn}).

\subsection{Graph Convolutional Networks}
GCN \cite{kipf2016semi} applies a first-order aggregation on graphs using affinity matrices. GCN contains several hidden layers that take a feature matrix $H^{(l)} \in R^{n\times d_l}$ as the input of the $l$-th layer and update the feature matrix $H^{(l+1)} \in R^{n\times d_{l+1}}$ by using a graph convolution operator. 

Given an input feature matrix $H^{(0)} \in R^{n\times d_0}$ and the graph affinity matrix $A\in R^{n\times n}$, GCN conducts the following layer-wise propagation,
\begin{equation} \label{eq:gcn}
    H^{(l+1)} = \sigma((I+S)H^{(l)}W^{(l)}),
\end{equation}
where $l \in \{ 1,2,...,L\}$ is the layer index and S is the same as in Eq.\ref{eq:s}. $W^{l}\in R^{d_l\times d_{l+1}}$ is the trainable weight matrix of the $l$-th layer. $\sigma(\cdot)$ denotes an activation function. Eq.\ref{eq:gcn} adopts $S$ rather than $A$ because multiplication with $A$ will completely change the scale of the feature vectors, which can cause numerical instabilities. To further alleviate this issue, \textcite{kipf2016semi} suggested to use the re-normalized $(I+S)$ as follows:
\begin{equation}
    I + D^{-1/2}AD^{-1/2} \rightarrow \hat{D}^{-1/2}\hat{A}\hat{D}^{-1/2},
\end{equation}
where $\hat{D}$ is the diagonal degree matrix of $\hat{A} = A + I$.

\begin{figure*}[ht]
\centering
\includegraphics[width=\textwidth]{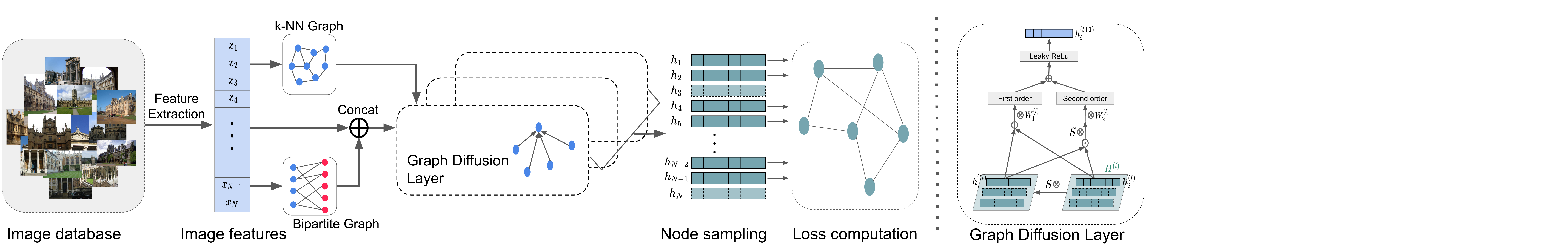}
\caption{{\bf The framework of GRAD-Net.} Given the input image features, GRAD-Net first constructs local sparsified k-NN graph and bipartite graph for the image instances. GRAD-Net refines the image features by the stacked graph diffusion layers that are optimized by global and local loss functions. The right panel is a detailed structure of the graph diffusion layer in GRAD-Net.}
\label{fig:pipline}
\end{figure*}

\subsection{Graph Diffusion Networks (GRAD-Net)} \label{sec:gdn}
In this section, we present our proposed GRAD-Net, illustrated in Fig. \ref{fig:pipline}, that is composed of multiple graph diffusion layers. These layers aggregate messages between nodes according to the manifold structure and output the updated node features. Lastly, an affinity estimation module takes the output features and estimates affinities. Apart from the local manifold structure information, a sparse coding feature vector $Z$ is also passed to the graph diffusion layer to provide global structure information. GRAD-Net is trained with two novel unsupervised loss functions, which we will discuss in details in Section \ref{sec:loss}. The details of the proposed GRAD-Net are as follows:

\subsubsection{Graph Constructions}
We first introduce how to construct graphs as the input to GRAD-Net. To encode both local and global structures from the image manifold, we design two types of graphs: local sparsified graph and global bipartite graph. 

\noindent \textbf{Local Sparsified Graph.} To construct the graphs of locality, we follow the mutual k-NN method in \textcite{iscen2017efficient} that computes the k nearest neighbors of every instance. The dataset is interpreted as a weighted graph $G=(V,E)$, consisting of $N=n+m$ nodes $v_i \in V$, and the edges $e_{ij} \in E$ connect those node pairs that have a non-zero affinity value, for $i,j\in \{1,2,...,N\}$. The affinity matrix $A=[a_{ij}] \in R^{N \times N}$ is defined as
\begin{equation} \label{eq:aij}
a_{ij}=\begin{cases}
			s(\bar{\mathcal{X}}_i,\bar{\mathcal{X}}_j), & \text{if  } \bar{\mathcal{X}}_i\in NN_k(\bar{\mathcal{X}}_j), \bar{\mathcal{X}}_j\in NN_k(\bar{\mathcal{X}}_i)\\
            0, & \text{otherwise}
		 \end{cases}
	,
\end{equation}
where $NN_k(\bar{\mathcal{X}}_i)$ denotes the set of $k$ nearest neighbors of node $i$, and $s: R^d\times R^d \rightarrow R$ is a predefined similarity metric typically based on cosine similarity or Euclidean distance. Eq.\ref{eq:aij} enables $A$ to be sparse when k is small, which makes diffusion on large graphs feasible. The edge weight is initialized as $a_{ij}$, and then the diffusion processes propagate the affinity values through the entire graph.

\noindent \textbf{Global Bipartite Graph by Sparse Coding.} To efficiently encode the global structure, we adopt a similar sparse coding in \textcite{liu2010large}, in which we select anchor nodes to represent the original data distribution and reconstruct the original nodes. 

We first apply a simple clustering algorithm, K-means, to select $B$ anchor nodes, and denote the feature matrix of these anchor nodes as $U\in\mathcal{R}^{B\times d}$. We then construct a bipartite graph $G_g = \{V_g,E_g\}$, where $V_g$ is the union of the original image nodes and the selected anchor nodes (i.e. $|V_g| = N + B$), and $E_g$ denotes all the edges between original images nodes and anchor nodes (i.e. $E_g \in \mathcal{R}^{N\times B}$). Let $Z \in \mathcal{R}^{N\times B}$ denote all the weights on the edge set $E_g$. We reconstruct each node by learning the edge weights, $\mathbf{z}_i$, that connect $\mathbf{x}_i$ to its $c$ closest neighboring anchor nodes by
\begin{equation}
\min_{\mathbf{z}_i} \|\mathbf{x}_i - U\mathbf{z}_i\|_2^2, \: \: s.t., \mathbf{\ell}^T\mathbf{z}_i = 1, \mathbf{z}_i \ge 0, |\mathbf{z}_i| = c.
\label{eq:sparsecoding}
\end{equation}

The optimization of Eq.\ref{eq:sparsecoding} is a typical sparse coding problem, which we adopt the same technique as in \textcite{lan2017learning}. We then obtain the global bipartite graph with edge weights $Z \in \mathcal{R}^{N\times B}$, and concatenate $Z$ as additional features to the image descriptors $\bar{\mathcal{X}}$.

\subsubsection{Graph Diffusion Layers.} 

The input graph $G$ consists of an affinity matrix $A\in R^{N\times N}$ and instance features $\bar{\mathcal{X}}$ as in Eq.\ref{eq:input}. We denote the input feature matrix $H^{(0)}\in R^{N\times d}$ in the first layer as,
\begin{equation}
    H^{(0)} = [\bar{\mathcal{X}}_1, \bar{\mathcal{X}}_2,...,\bar{\mathcal{X}}_N]^T
\end{equation}
The graph diffusion layer of GRAD-Net has a layer function $f_d$ as follows,
\begin{equation}
    H^{(l+1)} = f_d(H^{(l)}, A; W^{(l)}),
\end{equation}
where $W^{(l)}$ is the trainable parameters of the $l$-th layer, and $H^{(l)} \in R^{n\times d_l}$ and $H^{(l+1)} \in R^{n\times d_{l+1}}$ are the input and output features of the $l$-th layer, respectively. $d_l$ denotes the dimensions of the features in $l$-th layer. The total layer function $f_d$ consists of a first-order operator and a second-order operator.

The first-order operator, $f_1$, is a function that generates an updating message for node $u$ based on its first-order neighbor node $i$. The first-order message generated by $f_1$ is, 
\begin{equation}
    m^{(l)}_{u\leftarrow i} = f^{(l)}_1(h^{(l)}_i, p^{(l)}_{u\leftarrow i}),
\end{equation}
where $h^{(l)}_i$ denotes the feature of node $i$ in the $l$-th  layer, and $p^{(l)}_{u\leftarrow i}$ controls the scale of this message. We employ a similar first-order transition operation as in the conventional diffusion (i.e. $Sf^t$ in Eq.\ref{eq:rw}) in our graph diffusion layer, and incorporate additional trainable parameters $W_1$. In the $l$-th layer, $f_1$ is defined as,
\begin{equation} \label{eq:f1}
    f_1^{(l)} = SH^{(l)}W_1^{(l)},
\end{equation}
where $W_1^{(l)} \in R^{d_l\times d_{l+1}}$. This function turns out to share a similar form as the GCN layer in Eq.\ref{eq:gcn}.

The second-order operator, $f_2$, is a function that takes in the information of node $j$, a second-order neighbor of node $u$ (i.e. the neighbor of the neighbor of $u$), to update the feature vector of node $u$. The second-order message generated by $f_2$ is,
\begin{equation}
    m^{(l)}_{u\leftarrow \leftarrow j} = f^{(l)}_2(h^{(l)}_j, p^{(l)}_{u\leftarrow \leftarrow j}),
\end{equation}
where $h^{(l)}_j$ is the feature vector of node $j$ at the $l$-th layer, and $p^{(l)}_{u\leftarrow \leftarrow j}$ controls the scale of this message. The double arrow, $_{\leftarrow\leftarrow}$, represents the second-order connectivity between node $j$ and node $u$.

The challenge of designing $p^{(l)}_{u\leftarrow \leftarrow j}$ lies in the fact that node $j$ and $u$ are not adjacent, i.e. $a_{ju} = 0$. Inspired by the TPG diffusion of Eq.\ref{eq:tp} \cite{yang2012affinity}, we introduce $f_2$ as
\begin{equation}
   f_2^{(l)} = S(SH^{(l)}\odot H^{(l)})W_2^{(l)},
\end{equation}
where $\odot$ denotes the element-wise product, and $W_2^{(l)} \in R^{d_l\times d_{l+1}}$ is the trainable parameters at the $l$-th layer. The second-order operator, $f_2^{(l)}$, can be interpreted as a two-step hop. Let node $i$ be the shared neighbor of node $j$ and $u$. The product $SH^{(l)}$ is a one-step of message passing that resembles the first-order operation in Eq.\ref{eq:f1}, i.e. $m^{(l)}_{i\leftarrow j}$. Then $m^{(l)}_{i\leftarrow j} \odot H^{(l)}$ can be considered as the weighted feature of node $i$.
Lastly, $S(m^{(l)}_{i\leftarrow j}\odot H^{(l)})W_2^{(l)}$ passes the message from node $j$ to $u$ through their common shared neighbor node $i$. 
This second-order propagation enlarges the representational power of the graph diffusion layer.

Collectively, the graph diffusion layer performs the propagation as
\begin{equation} \label{eq:gdn}
\begin{aligned}
H^{(l+1)} & = f_d(H^{(l)}, A; W^{(l)}) = \sigma (m^{(l)}_{u\leftarrow i} + m^{(l)}_{u\leftarrow \leftarrow j}) \\
& =\sigma((I+S)H^{(l)}W_1^{(l)}+S(SH^{(l)}\odot H^{(l)})W_2^{(l)}),
\end{aligned}
\end{equation}
where $\sigma$ is an activation function and is set to LeakyRelu \cite{xu2015empirical} in GRAD-Net, and $W_1^{(l)}, W_2^{(l)} \in R^{d_l\times d_{l+1}}$ are the trainable parameters.
An identity matrix, $I$, is added to the first order operation to introduce a self-loop propagation. The learned features of GRAD-Net can be the output of the last layer, $H^{(L)}$, or the concatenated layer features, $H$, 
\begin{equation} \label{eq:H}
    H = H^{(0)}\| H^{(1)}\|... \|H^{(L)},
\end{equation}
where $\|$ is the concatenation along the feature dimension.

\noindent \textbf{Affinity Estimation.} Pairwise similarity is calculated using the concatenated layer features, $H\in R^{N \times D}$ (Eq.\ref{eq:H}). GRAD-Net adopts cosine similarity. The similarity between node $i$ and node $j$, $s_{ij}$, is
\begin{equation} \label{eq:sim}
    s_{ij} = cosine\ similarity(h_i, h_j)=\frac{h_i\cdot h_j}{\|h_i\|\|h_j\|},
\end{equation}
where $h_i\in R^D$ and $h_j\in R^D$ are the $i$-th and $j$-th rows of $H$, respectively.

\subsection{Loss Functions for Global and Local Consistency} \label{sec:loss}
Loss functions are crucial for training the GRAD-Net model.
To capture both local and global structures of the underlying data manifold, we propose two loss functions to optimize \textit{local smoothness} and \textit{global order}, which will be illustrated in this section.

\begin{figure}[h]
\centering
\includegraphics[width=0.45\textwidth]{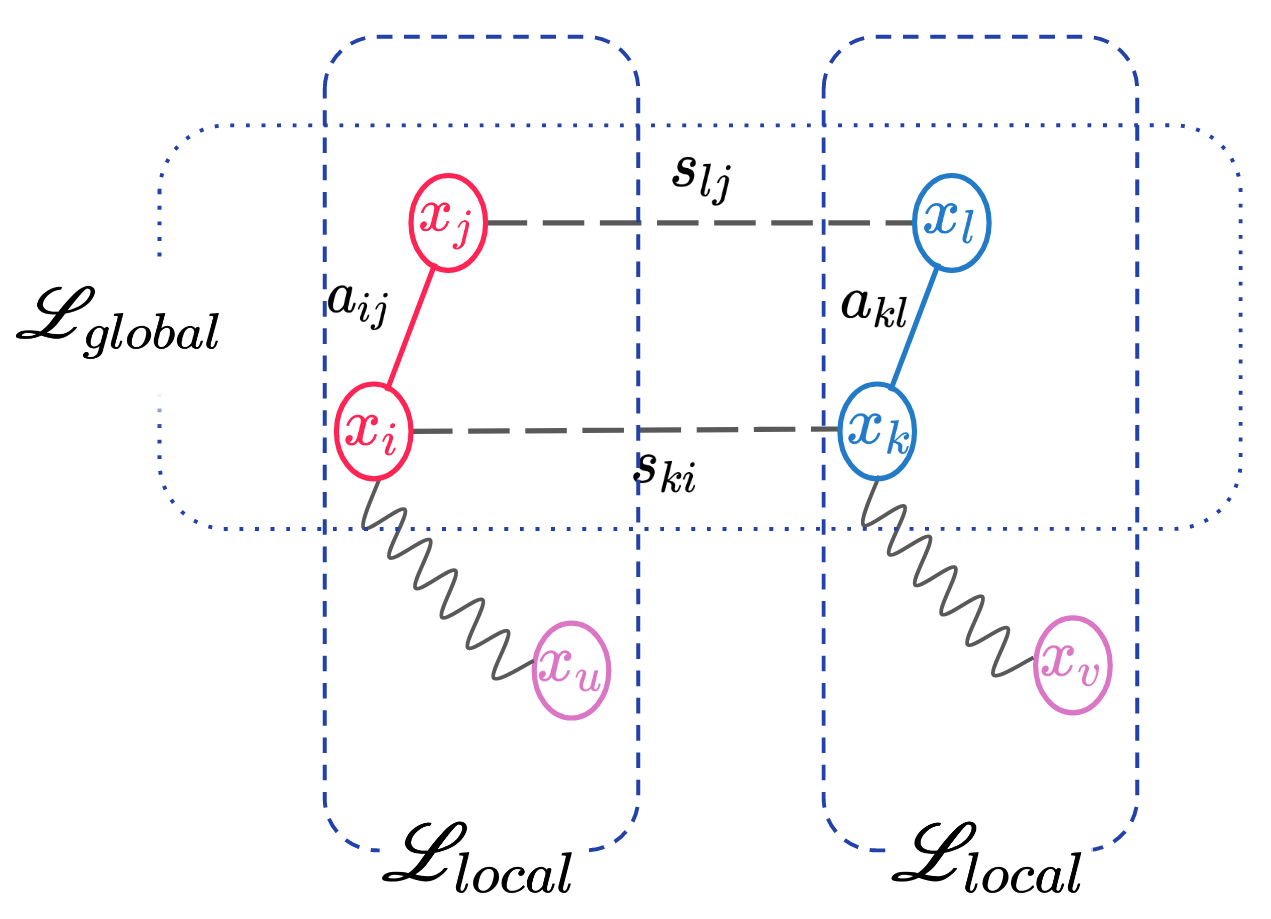}
\caption{{\bf Illustration of sextet.} (i) Triplet node $(u, i, j)$: $x_i$ and $x_j$ are first-order neighbors, and $x_u$ is a non-neighbor of $x_i$ on the graph. (ii) Quadruplet nodes $(i,j,k,l)$: $x_i$ and $x_j$ are first-order neighbors and $x_l$ and $x_k$ are first-order neighbors. (iii) Sextet $(i,j,k,l,u,v)$: the collection of $x_i$, $x_j$, $x_k$, $x_l$, $x_u$ and $x_v$. \textbf{Global order.} $s_{lj}$ and $s_{ki}$ both measure the similarities between two neighborhoods and thus should be similar. $\mathscr{L}_{local}$ takes in the triplet nodes $(u, i, j)$. $\mathscr{L}_{global}$ takes in the quadruplet nodes $(i,j,k,l)$.}
\label{fig:loss}
\end{figure}

\noindent \textbf{Local smoothness} refers to the assumption that if two nodes are topologically closer in the graph $G$, the similarity of their features should be higher. %
To regularize local smoothness, we propose a similar loss as the pairwise Bayesian Personalized Ranking (BPR) loss \cite{rendle2009bpr},
\begin{equation}\label{eq:lgo}
    \mathscr{L}_{local} = \sum_{(i,j,u)\in \mathscr{O}_{\mbox{\tiny \it local}}}-ln(s_{ij}-s_{iu}),
\end{equation}
where $s_{ij}$ is the similarity between the features of instances $i,j$, $\mathscr{O}_{\mbox{\tiny \it local}} = \{(i,j,u)|j\in \mathscr{N}_i, u\notin \mathscr{N}_i \}$ denotes the triplet training samples, and $\mathscr{N}_i$ refers to the nodes that share direct edge connections with $i$ in the graph $G$.

The local smoothness loss in Eq.\ref{eq:lgo} can be interpreted as the difference of the similarities between a close node-pair and a distant node-pair. The local loss function enforces larger similarities between the nodes in the same local neighborhood, which is consistent with the empirical findings that the diffusion process should be performed locally on the manifold \cite{wang2012affinity, donoser2013diffusion}.

\noindent \textbf{Global Order} refers to the assumption that the similarities measured by different nodes from two neighborhoods should remain consistent \cite{bai2018regularized,yang2012affinity}. As illustrated in Fig. \ref{fig:loss}, if node $i$ and $j$ are from one neighborhood while node $k$ and $l$ are from another neighborhood, both $s_{lj}$ and $s_{ki}$ should reflect the similarities between the two neighborhoods, and thus should be similar under the global order assumption. To enforce global order, we introduce a loss function suitable for neural networks training,
\begin{equation} \label{eq:lls}
    \mathscr{L}_{global} = \sum_{(i,j,k,l)\in \mathscr{O}_{\mbox{\tiny \it global}}} ln(1+\beta \cdot a_{ij}a_{kl}s_{ki}s_{li}(s_{ki}-s_{lj})^2),
\end{equation}
where $\mathscr{O}_{\mbox{\tiny \it global}}=\{ (i,j,k,l)|j\in\mathscr{N}_i,j\in \mathscr{N}_k \}$, $\beta \in R^+$ is a weighting coefficient, $a_{ij}$ is the affinity in the manifold affinity matrix $A$, and $s_{ki}$ is the similarity defined in Eq.\ref{eq:sim}.  %

The total loss function is a weighted combination of the local and global loss functions:
\begin{equation} \label{eq:total_l}
    \mathscr{L} =  \mathscr{L}_{local} + \alpha \mathscr{L}_{global} + \lambda \|\Theta\|^2_2,
\end{equation}
where $\Theta$ is the set of all the trainable parameters and $\| \cdot \|^2_2$ is the l2-norm function. $\alpha \in R^+$ and $\lambda\in R^+$ are the weights of different loss components.

The local loss in Eq.\ref{eq:lgo} works on triplet nodes $(u,i,j)$, while the global loss in Eq.\ref{eq:lls} works on quadruplet nodes $(i,j,k,l)$. Considering this, we provide a compact solution to implement the training process. For every training iteration, a sextet $(i,j,k,l,u,v)$ is selected to compute the total loss (Eq.\ref{eq:total_l}) as shown in Fig. \ref{fig:loss}, where the local loss $\mathscr{L}_{local}$ is calculated twice from the two triplets $(i,j,u)$ and $(k,l,v)$, while the global loss $\mathscr{L}_{global}$ is calculated once from the quadruplet $(i,j,k,l)$. Therefore, the implemented loss function is
\begin{equation} \label{eq:actual_l}
    \begin{split}
    \mathscr{L}(i,&j,k,l,u,v) \\
    &= \frac{1}{N_{batch}}\sum_{batch} \{\mathscr{L}_{local}(i,j,u) + \mathscr{L}_{local}(k,l,v) \\
    &+ \alpha \mathscr{L}_{global}(i,j,k,l) + \lambda \|\Theta\|^2_2 \},
    \end{split}
\end{equation}
where $N_{batch}$ is the number of sextets in a mini batch.

\subsection{Inductive Learning on Unseen Instances} \label{sec:ilui}
Most existing methods assume the queries are observed in the dataset, yet this setting cannot be met in many real-life scenarios. When queries are only accessible in the production stage after deployment, conventional diffusion methods suffer substantially from computation burden, and thus lack the feasibility to process queries in real-time. One solution to this issue is Query Expansion (QE) \cite{chum2007total}, in which a new query is constructed by averaging the image features of its top nearest neighbors in the original database.

We extend QE to derive a fast query expansion approach, the query feature expansion (QFE), that is compatible with our feature learning fashion. 
Given a new query $q$, QFE first finds the $k$ nearest neighbor $\mathscr{N}_{q}$ of $q$ using original image descriptors, and directly computes the features of $q$ in the learned feature space by 
\begin{equation} \label{eq:qfe}
    h_q =  \sum_{i\in \mathscr{N}_{q}}s_{qi}h_i,
\end{equation}
where $h_i$ is the learned feature of node $i$, and $s_{qi}$ is the similarity between $q$ and its neighbor $i$ computed by the original image descriptors. $h_q$ is a feature approximation for $q$ and can be used to retrieve instances via similarity search. We demonstrate the effectiveness of QFE with empirical results in Section \ref{sec:il}.

\subsection{Training on Large Datasets} \label{sec:large}
\noindent \textbf{Mini-batch Training on Graph.}
We found out that only the nodes in the computation flow of the loss function (Eq.\ref{eq:total_l}) will contribute to the parameter updating process. 

To alleviate the computation burden on large-scale datasets, we adopt a mini-batch approach similar to Breadth-First Search (BFS) \cite{hamilton2017inductive} on the sparsified mutual k-NN graph. We start from a batch of sextets ($N_{batch} \times 6$ nodes) to find $N'_{batch}$ relevant nodes, and then only include these $N'_{batch}$ of nodes in each training iteration. This acceleration process remarkably speeds up by over 20 times on training large-scale datasets (e.g. Oxford105k \cite{iscen2017efficient}). 

\noindent \textbf{Truncation.} As image retrieval on very large datasets is challenging to diffusion-based methods, truncation has become a common practice to alleviate this issue \cite{iscen2017efficient, bai2018regularized}. 
Though our proposed GRAD-Net is capable of training and retrieving datasets with over 100k instances efficiently, we implement a similar truncation method to further accelerate the training process.
GRAD-Net first forms a union graph of the top 500 nearest neighbors for each query, and uses the learned features based on this union graph for the retrieval task. 
We demonstrate the efficacy of mini-batch training and truncation on various scales of datasets in Fig. \ref{fig:complexity}. 

\subsection{Implementation Details} \label{sec:imp}
In this section, we address several implementation considerations, and introduce the default configurations of GRAD-Net. For all the experiments in Section \ref{sec:exp} and \ref{sec:disc}, we deploy the default configurations unless otherwise specified. 

We implemented GRAD-Net on the PyTorch\cite{paszke2019pytorch} framework. All the experiments were conducted on a workstation with a 12-core Intel(R) Xeon(R) CPU W-2133 @ 3.60GHz and one NVIDIA RTX5000 GPU with 16GB GPU memory.

The number of neighbors, $k$, in the mutual k-NN search is set to 15. We set the number of anchor points $B=100$ to compute the sparse features $Z$. The similarity metric $s$ in Eq.\ref{eq:aij} is set to $1/(1+Euclidean\ Distance)$. The number of graph diffusion layers, $L=3$, with $d_1 = 1024$, $d_2 = 256$ and $d_3 = 128$. An l2-normalization followed by dropout \cite{srivastava2014dropout} with a probability of 0.3 is applied to the output of each layer. We set $\beta=e^{5}$ in Eq.\ref{eq:lls}, $\alpha=1.0$ and $\lambda=e^{-5}$ in the total loss function (Eq.\ref{eq:actual_l}). The sextet $(i,j,k,l,u,v)$ are sampled as follows: node $i,k$ are first randomly sampled from all nodes, then nodes $j, l$ are randomly selected from the first-order neighbors of node $i$ and $k$, and lastly nodes $u,v$ are randomly sampled from the non-neighbor nodes of $i, j, k, l$. We use Adam \cite{kingma2014adam} to optimize the parameters. The learning rate of Adam is initialized as $3 \times e^{-4}$ and reduced by $50\%$ at 30 and 100 epochs, respectively. The model is trained for 300 epochs. The training loss reaches a stable state after 200 epochs, and thus we take the model at the 300-th epoch as the final model for evaluation. The number of sextets in a mini-batch, $N_{batch} = 64$. %
We also incorporate the efficient similarity search library, faiss \cite{johnson2019billion}, to boost up our mutual k-NN search and employ inverted index to reduce the memory consumption of the dataset instances. We store the highly compressed faiss index instead of the original representations in deployment. %

\section{Experiment} \label{sec:exp}
To demonstrate the performance of GRAD-Net, we conducted experiments on a synthetic toy dataset and seven popular benchmarking datasets, including %
face images (Section \ref{sec:face}) and natural images (Sections \ref{sec:oxford}, \ref{sec:paris}, \ref{sec:roxford}).
Table \ref{tb:ds} summarizes the type and statistics of all the analyzed datasets. All the implementation settings follow the descriptions in Section \ref{sec:imp}. %

\begin{table}[h]
\renewcommand{\arraystretch}{1.3}
\caption{Summary of datasets}
\label{tb:ds}
\centering
\begin{tabular}{|c|c|c|c|c|}
\hline 
Dataset & Type & Classes & Instances & Feature Dims\\
\hline
ORL\cite{samaria1994parameterisation} & Face & 40 & 400 & 10,304 \\
Oxford5k\cite{philbin2007object} & Image & 11 & 5,062 & 512 or 2,048 \\
Oxford105k\cite{philbin2008lost}& Image & 11 & 105K & 512 or 2,048 \\
Paris6k\cite{philbin2007object} & Image  & 11  & 6,392 & 512 or 2,048  \\
Paris106k\cite{philbin2008lost}& Image & 11 & 106K & 512 or 2,048  \\
$\mathcal{R}$Oxford\cite{radenovic2018revisiting} & Image  & 11  & 4,993 & 2,048  \\
$\mathcal{R}$Paris\cite{radenovic2018revisiting} & Image  & 11   & 6,322 & 2,048  \\
\hline
\end{tabular}
\end{table}
\subsection{Toy example}

\begin{figure*}[h!]
\centering
\includegraphics[width=1.0\textwidth]{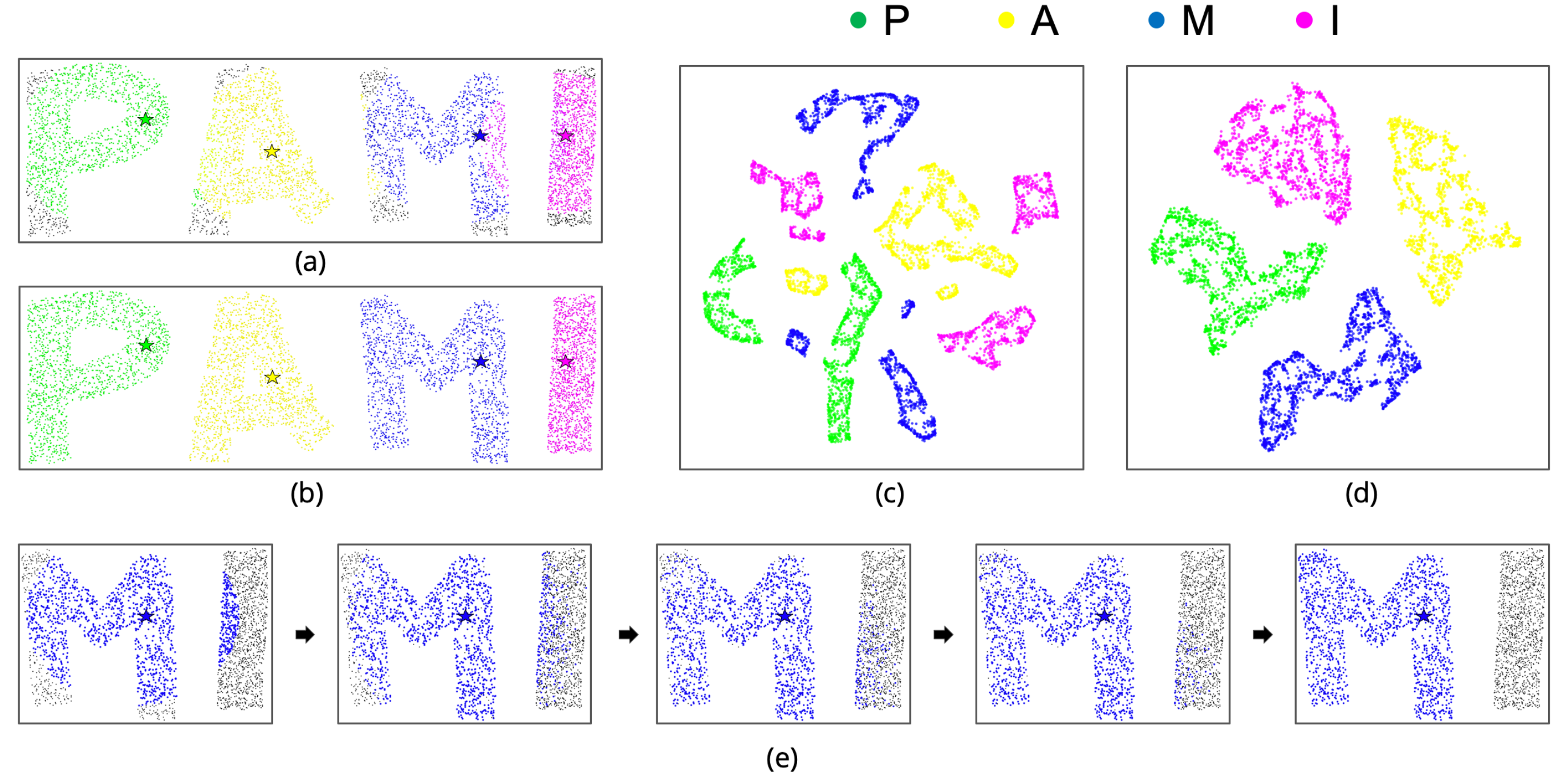}
\caption{{\bf Synthetic toy dataset results.} (a) Retrieval results based on Euclidean distance as similarity metric; (b) retrieval results using GRAD-Net; (c) t-SNE visualization of the original features; (d) t-SNE visualization of the features output by GRAD-Net; (e) initial, intermediate and final retrieval results of the blue query point using GRAD-Net.}
\label{fig:pami}
\end{figure*}

We generated 1,500 points in 2-D space that construct the word \textit{PAMI} to demonstrate that GRAD-Net can well capture the geometric manifold. We take the coordinates of these points as the original features. The query points are chosen in each letter (marked by star) colored in green, yellow, blue, and purple, respectively. 

Fig. \ref{fig:pami}(a) uses Euclidean distance as the similarity metric to find the $k$ nearest neighbors of each query point while Fig. \ref{fig:pami}(b) uses GRAD-Net to retrieve the relevant points for each query. Take the blue query point in the letter $M$ for example. Fig. \ref{fig:pami}(e) depicts the initial, intermediate and final retrieval results using GRAD-Net. At the initial step, some points in the letter $I$ that are closer to $M$ by the Euclidean metric and were wrongly retrieved to the blue query point. However, as the training progresses, GRAD-Net is capable of correctly retrieving points from $M$ for the blue query point. 

Fig. \ref{fig:pami}(c) shows the t-SNE plot \cite{maaten2008visualizing} of the original features, while Fig. \ref{fig:pami}(d) shows the t-SNE plot of the features produced by GRAD-Net. The t-SNE plots clearly verify that GRAD-Net better captures the intrinsic manifold than Euclidean distances.

\subsection{ORL Dataset} \label{sec:face}
\begin{table*}[h]
\renewcommand{\arraystretch}{1.3}
\caption{Performance comparison (bullseye score) on the ORL dataset.}
\label{tb:orl}
\centering
\begin{tabular}{|c|c|c|c|c|c|c|c|c|c|c|c|}
\hline
Methods & k-NN & SD \cite{wang2012affinity} & LCDP \cite{yang2009locally}& TPG \cite{yang2012affinity} & MR \cite{zhou2004ranking} & GDP \cite{donoser2013diffusion} &  RDP \cite{bai2018regularized} & Efficient \cite{yang2019efficient} & GSS \cite{liu2019guided}& GRAD-Net(Ours) \\
\hline
Bullseye score & 62.35 & 71.67 & 74.25 & 73.90 & 77.58 & 77.42 & 79.27 &82.00* &83.87* & \textbf{84.14} \\
\hline
\end{tabular}
\begin{flushleft}
$^{*}$Scores are reported by using the code provided by the authors.
\end{flushleft}
\end{table*}
The ORL dataset \cite{samaria1994parameterisation} is a face image dataset that contains 400 images of size $112 \times 92$ taken from 40 distinct subjects. The images vary in illuminations, facial expressions and facial details. We first vectorize and normalize the raw image pixels as image descriptors. 

Retrieval accuracy is measured by the average recall rate at a window size $K$ for each query, also known as the bullseye score. For this dataset, we set $K=15$ and the baseline bullseye score is $62.35\%$. Table \ref{tb:orl} summarizes the performance of various approaches on this dataset. Our proposed GRAD-Net achieves the state-of-the-art performance.

\begin{table*}[t!]
\renewcommand{\arraystretch}{1.3}
\caption{Performance comparison (mAP scores) on Oxford5k, Oxford105k, Paris6k, Paris106k, $\mathcal{R}$Oxford and $\mathcal{R}$Paris}
\label{tb:oxford}
    \centering
        \begin{tabular}{|c|c|c|c|c|c|c|c|}
        \hline
        Method & Descriptor & Oxford5k & Oxford105k & Paris6k & Paris106k & $\mathcal{R}$Oxford & $\mathcal{R}$Paris\\
        \hline
        k-NN & \multirow{6}{2em}{VGG} & 79.5 & 72.1 & 84.5 & 77.1 & \multirow{6}{1em}{-} & \multirow{6}{1em}{-} \\

        k-NN + AQE \cite{chum2007total} & & 85.4 & 79.7 & 88.4 & 83.5 & &\\
        Regional Diffusion \cite{iscen2017efficient} & & 85.7 & 82.7 & 94.1 & \textbf{92.5} & &\\
        Efficient \cite{yang2019efficient} & & 88.2 & \textbf{85.0} & 94.7 & 92.2 & & \\
        GSS \cite{liu2019guided} & & 87.8 & OOM & 93.7 & OOM & &\\
        \textbf{GRAD-Net(Ours)} & & \textbf{90.1} & 84.6 & \textbf{94.8} & 91.2 & &\\
        \hline
        k-NN & \multirow{6}{2em}{ResNet} & 83.6 & 80.8 & 93.8 & 89.9 & 64.7 & 77.2 \\
        k-NN + AQE \cite{chum2007total} & & 89.6 & 88.3 & 95.3 & 92.7 & 67.2 & 80.7 \\
        Regional Diffusion \cite{iscen2017efficient} & & 87.1 & 87.4 & 96.5 & 95.4 & 69.8 & 88.9\\
        Efficient \cite{yang2019efficient} & & 92.6 & 90.8 & 97.1 & 94.9 &72.1& 91.3\\
        GSS \cite{liu2019guided} & & 91.5 & OOM & 96.1 & OOM & \textbf{77.8} & \textbf{92.4} \\
        \textbf{GRAD-Net(Ours)} & & \textbf{95.9} & \textbf{94.5} & \textbf{97.3} &\textbf{95.4} & 75.6 & 90.6\\
        \hline
        \end{tabular}
\begin{flushleft}
\hspace{1.8cm} (i) OOM stands for out of memory. \\
\hspace{1.8cm} (ii) VGG style descriptors are not available for the $\mathcal{R}$Oxford and $\mathcal{R}$Paris datasets. \\
\hspace{1.8cm} (iii) Scores of Efficient and GSS are reported by using the code provided by the authors.
\end{flushleft}
\end{table*}

\subsection{Oxford5k and Paris6k Dataset} \label{sec:oxford}
In this section, we examine the performance of GRAD-Net with real natural images on the Oxford5k and Paris6k datasets \cite{philbin2007object}. 

\noindent \textbf{Experiments on Oxford5k.} The Oxford5k dataset consists of 5,062 pictures of 11 different Oxford buildings collected from Flickr and 55 query images with the ground truth for evaluation.

For a fair comparison, we employed the image descriptors provided in \textcite{iscen2017efficient} to perform the experiments. In particular, the image descriptors are in two types: the first type consists of 512 dimensional descriptors \cite{radenovic2016cnn} derived from a fine-tuned VGG net, while the second type consists of 2,048 dimensional descriptors \cite{gordo2017end} derived from a fine-tuned ResNet. For any given instance, \textcite{iscen2017efficient} extracts one global feature vector and multiple regional feature vectors. In this experiment, we only consider the scenario where an instance is represented by one global feature vector. Therefore, we only compare GRAD-Net to the models using the global features as input here and after. 

\begin{figure}[h!]
\centering
\includegraphics[width=0.49\textwidth]{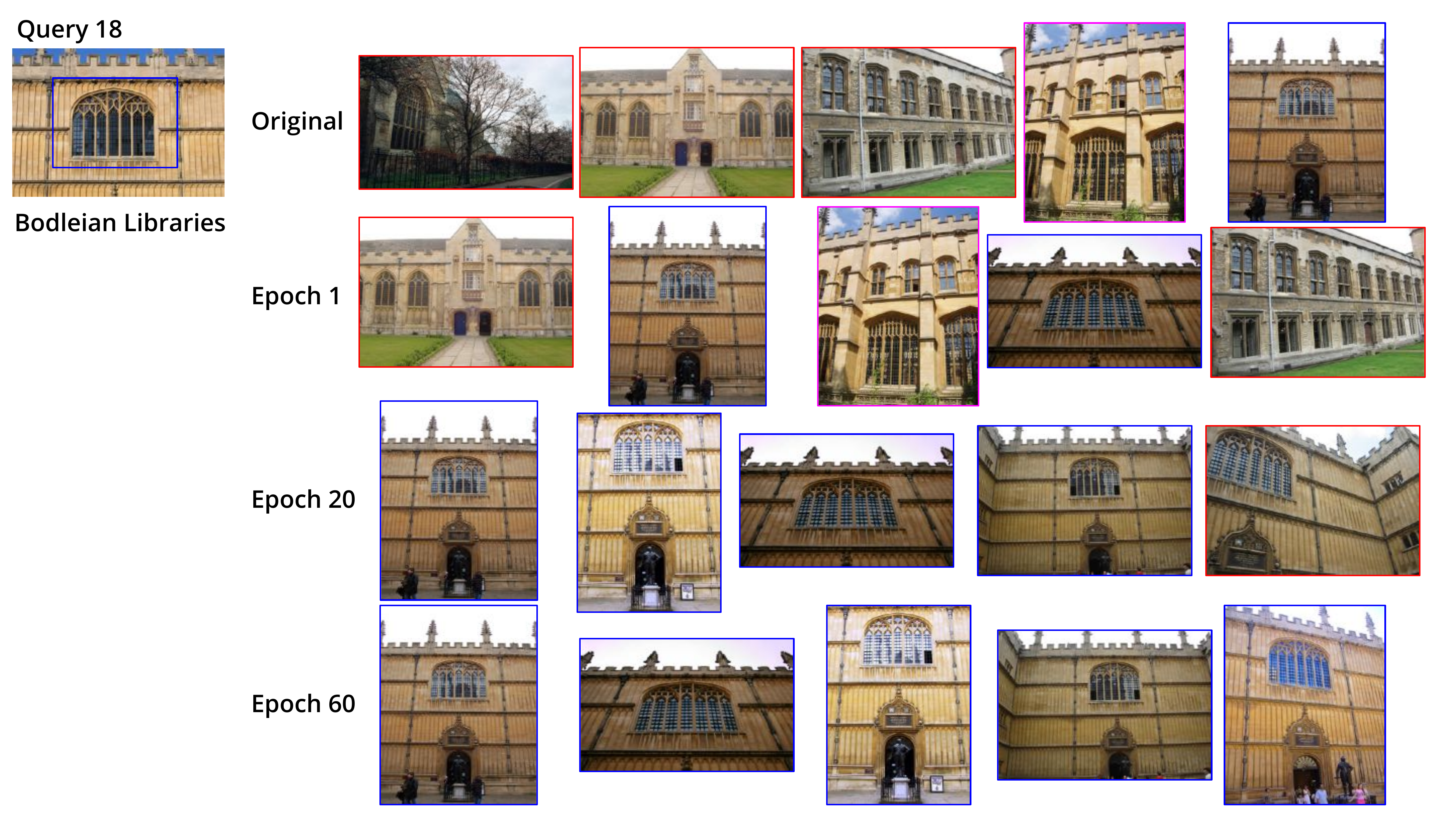}
\caption{{\bf Retrieval results on the Oxford5k dataset at different epochs.} The query instance is on the top left. Each row on the right contains the top five retrieved images after the corresponding training epoch. Images in the red frames are the incorrect results, while images in the blue frames are the correct ones. As the training progresses, GRAD-Net is capable of retrieving the correct instances for the query.}
\label{fig:te}
\end{figure}

We compare GRAD-Net with existing methods and summarizes the results in Table \ref{tb:oxford}. We use the standard evaluation protocol, mean Average Precision (mAP) ranging from $0\%$ to $100\%$, to quantify the retrieval accuracy. For both types of image descriptors, GRAD-Net significantly improves the baseline performances and achieves the highest mAP among all the methods, $90.1\%$ for the VGG descriptors and $95.9\%$ for the ResNet descriptors, exceeding the existing state-of-the-art method by a large margin. Fig. \ref{fig:te} depicts the retrieval results by GRAD-Net at the initial state, epoch 1, 20 and 60, respectively, which demonstrates that the training process improves the retrieval results of GRAD-Net. 

We also showed that a simple nearest neighbor search using the sparse vectors $Z$ achieves impressive mAP score of $87.5\%$ on the ResNet image descriptors. This demonstrates that our proposed sparse coding method is able to capture global similarity relations and reduce the noise in original dense vectors. GRAD-Net further improves the mAP score by additional $8.4\%$, which indicates that GRAD-Net is capable of recovering the manifold structure.

\noindent \textbf{Experiments on Paris6k.} The Paris6k dataset was collected in a similar fashion to the Oxford5k dataset. It includes 6,392 images of 11 buildings in Paris and uses 55 queries with the ground truth for evaluation. We also employed the VGG and ResNet features from \textcite{philbin2008lost} and adopted the same model configurations as in the Oxford5k experiment, except for $k=60$ for the mutual k-NN graph. Results in Table \ref{tb:oxford} show that GRAD-Net outperforms the existing state-of-the-art on both types of image descriptors.

\subsection{Oxford105k and Paris106k Datasets} \label{sec:paris}
To demonstrate the efficacy of GRAD-Net on large-scale datasets, we evaluate our proposed method on the Oxford105k and Paris106k datasets \cite{philbin2008lost}. These datasets are the extensions of the Oxford5k and Paris6k datasets with additional 100k irrelevant images from Flickr as distractors. 

Due to the large scale of the datasets, most conventional diffusion methods fail to process the entire dataset within an acceptable time. For example, the diffusion iteration for one query from Oxford105k takes 13.9s with a 12-core CPU \cite{iscen2017efficient}. On the contrary, GRAD-Net, deploying mini-batch training and feature truncation, is capable of handling the $>$100k instances in both datasets (see Section \ref{sec:ca} for the detailed complexity analyses).

As shown in Table \ref{tb:oxford}, GRAD-Net achieves outstanding performance (mAP score $94.5\%$) on the Oxford105k dataset using the ResNet descriptors and outperforms the existing state-of-the-art method by $2.7\%$. On the Paris106k dataset, GRAD-Net also achieves comparable performance to the state-of-the-art method (mAP score $95.4\%$ compared to $95.6\%$). Using the VGG descriptors, GRAD-Net also achieves comparable performance. We observed that the truncated union graphs generated from the VGG image descriptors omit some of the correct instances, which may partially explain the under-performance on the VGG descriptors.

\subsection{$\mathcal{R}$Oxford and $\mathcal{R}$Paris Datasets} \label{sec:roxford}
\textcite{radenovic2018revisiting} generated the $\mathcal{R}$Oxford and $\mathcal{R}$Paris datasets by manually examining the instances in Oxford5k and Paris6k, and selected a set of 70 challenging queries with ground truth for evaluation. We adopted the image descriptors from \textcite{radenovic2018fine} and used the \textit{Medium} evaluation protocol.

\begin{table*}[h!]
\renewcommand{\arraystretch}{1.3}
\caption{Ablation study results of the variants in GRAD-Net.}
\label{tb:abl}
\centering
\begin{tabular}{|c|c|c|c|c|}
\hline
Category & Variants & Epochs mAP$92+$ & mAP & Diff \\ %
\hline
- & base & 9 & 95.95$\pm$0.26 & -\\
\hline
\multirow{3}{10em}{Loss (default Eq.\ref{eq:actual_l})} & w/o $\mathscr{L}_{local}$ & -  & \textbf{90.16}$\pm$0.39 & \textbf{5.79}\\
                        & w/o $\mathscr{L}_{global}$  & 12 & 95.47$\pm$\textbf{0.42} & 0.48\\  %
                        & w/o $\|\Theta\|^2_2$ & 8  & 95.64$\pm$0.39 & 0.31\\  %
\hline
\multirow{2}{10em}{Sparse coding (default $Z \|$original features) }
                        & w/o $Z$& 12 & 95.56$\pm$0.24 & 0.39\\
                        & w/o original features & - & \textbf{89.51$\pm$0.53} & \textbf{6.44} \\
\hline
\multirow{5}{10em}{Model structure (default 1024, 256, 128)} 
                        & Layers-1024,256,128,128 & 9 & 95.28$\pm$0.35 & 0.67 \\
                        & Layers-1024,256 & 9 & 95.55$\pm$0.39 & 0.40\\
                        & Layers-512,128,64 & 15 & 95.39$\pm$\textbf{0.64} & 0.56\\
                        & GCN w/o 2nd order& 8 & 94.99$\pm$0.52 & \textbf{0.96} \\ %
                        & w/o concatenation & \textbf{20} & 92.51 $\pm$1.02 & \textbf{3.44}\\
\hline
\multirow{5}{10em}{Manifold structure (default $k$=15)} 
                        & $k$=8 & \textbf{30} & 93.07$\pm$0.54 & 2.88\\
                        & $k$=20 & 12 & 95.21$\pm$0.48 & 0.74\\
                        & $k$=25 & 10 & 94.03$\pm$\textbf{0.67} & 1.92\\
                        & $k$=32 & 8 & 92.68$\pm$0.62 & 3.35\\
                        & $k$=64 & - & 90.37$\pm$0.47 & \textbf{5.58}\\
\hline
\multirow{2}{10em}{Similarity metric (default-Euclidean)} 
                        & Gaussian Euclidean\cite{yang2012affinity} & \textbf{30} & 94.36$\pm$0.58 & \textbf{1.59}\\
                        & Cosine & 28 & 94.63$\pm$\textbf{0.64} & 1.32 \\
\hline
\multirow{4}{10em}{Batch size(default batch=64)} 
                        & batch=32 & 10 & 95.37$\pm$\textbf{0.47} & 0.58 \\
                        & batch=128 & 15 & 95.40$\pm$0.27 & 0.55 \\
                        & batch=256 & 21 & 94.84$\pm$0.30 & 1.11 \\
                        & batch=1024 & \textbf{62} & 93.69$\pm$0.37 & \textbf{2.04} \\
\hline
\end{tabular}
\begin{flushleft}
\hspace{3.1cm} The bold entries are the ones differ the most from the base model in each category. 
\end{flushleft}
\end{table*}

\section{Discussion} \label{sec:disc}
\subsection{Ablation Study} \label{sec:abl}

In this section, we vary the configurations of several important settings in GRAD-Net to verify the effectiveness of the corresponding variants. For the purpose of illustration, we focus on the Oxford5k dataset, and set the base model as the one discussed in Section \ref{sec:oxford} that achieves an mAP score of $95.95\%$.

For fair comparisons, we ran six trials for each variant and evaluated the performance with three evaluation metrics:
\begin{enumerate}
    \item \textit{Epochs mAP$92+$} refers to the number of epochs that a model takes to first reach an mAP score of $92\%$ during the training process. We use this metric to examine the learning efficiency.
    \item \textit{mAP} is the mean Average Precision score. We report the average and standard deviation of the mAP scores of the six trial runs. 
    \item \textit{Diff} refers to the mAP drop compared to the base model (the first row in Table \ref{tb:abl}). %
\end{enumerate}

\noindent \textbf{Loss functions.} The ablation study of the loss functions shows that $\mathscr{L}_{local}$ and $\mathscr{L}_{global}$ have substantial impact on the model performance. Without the local loss, the proposed model can only reach an average mAP score of $90.16\%$ and exhibits a slight increase of instability in the training process. The absence of the global loss, $\mathscr{L}_{global}$, also increases the standard deviation of the mAP score by $0.42\%$. We also tested the impact of global loss function on the Oxford105k dataset. The performance drops by $0.96\%$ without global loss. %

\noindent \textbf{Sparse coding.} The sparse coding vector $Z$ is another important component of GRAD-Net. By default, $Z$ is a 100-dimensional vector appended to the original features. We test on two variants related to $Z$: taking (i) original features only and (ii) $Z$ only as the inputs to GRAD-Net. The ablation studies show that both $Z$ and the original features are crucial to the retrieval performance.

\noindent \textbf{Model Structure.} We vary the layer settings of GRAD-Net, and the results show that GRAD-Net is robust to the layer size and the number of layers selection on the Oxford5k dataset. One interesting finding is that when replacing the graph diffusion layer (that includes a second-order operator) with the vanilla GCN network (that only includes a first-order operator), the mAP score drops by $0.96\%$. This finding empirically validates the effectiveness of the second-order operator in the diffusion layer of GRAD-Net. Moreover, we perform ablation studies on the GRAD-Net output features with or without the concatenation operation as in Eq.\ref{eq:H}. The results verify that the concatenation of features in Eq.\ref{eq:H} is essential and accounts for $3.44\%$ of the performance gain as well as decreases the variation of performance.

\noindent \textbf{Manifold Structure.} The test results on the manifold structures show that the value of $k$ in the mutual k-NN search is crucial to the learned features. On the Oxford5k dataset, $k$ is optimal around $15$, yet GRAD-Net demonstrates robust performance for $k$ ranging from $8$ to $25$. 

\noindent \textbf{Similarity metric and batch size.} The variants in the similarity metrics show that GRAD-Net is robust to different similarity metric choices. Experiments of various batch sizes show that smaller batch size is preferable for better performance.

\begin{figure}[h!]
\centering
\includegraphics[width=0.48\textwidth]{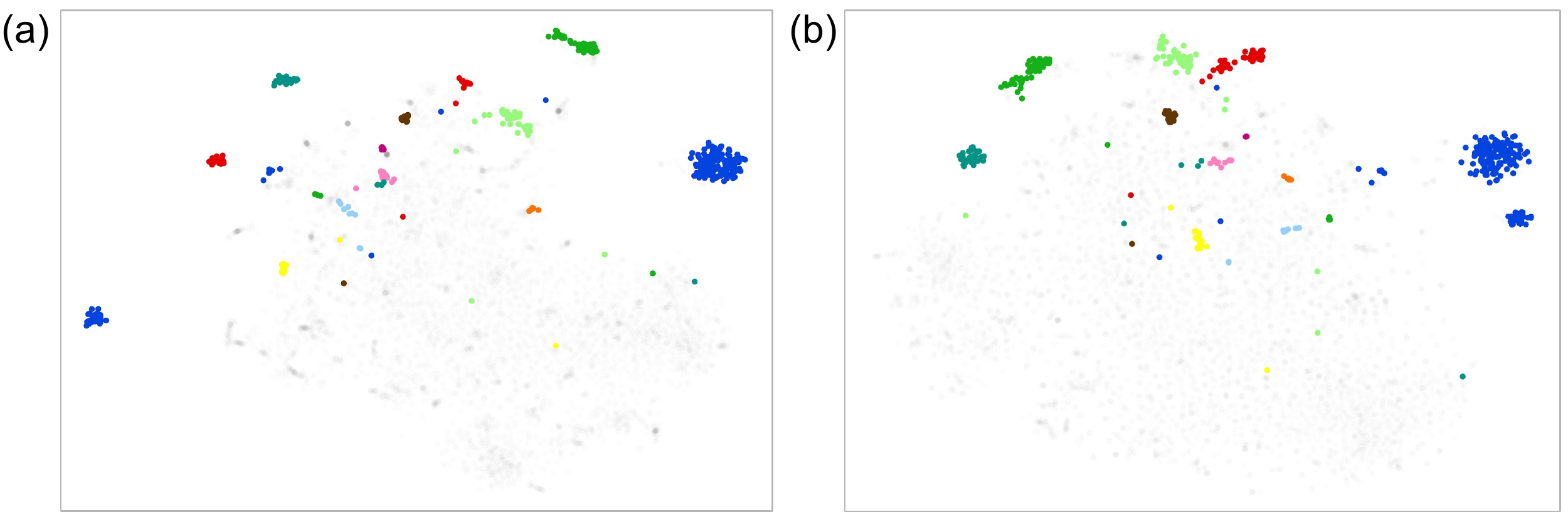}
\caption{{\bf The feature learning results.} (a) t-SNE visualization of the original image features; (b) t-SNE visualization of the learned features. Colors denote the image categories. Grey points are denote the images that do not belong to any query category.}
\label{fig:fl}
\end{figure}

\subsection{Feature Learning} \label{sec:fl}

The most noticeable difference between GRAD-Net and conventional diffusion methods is that GRAD-Net performs feature learning, while conventional diffusion methods only operates on the affinity matrix. 

Based on the results from Fig. \ref{fig:pami}(c-d), we argue that GRAD-Net learns a better representation of semantic information. We further validate the feature learning ability of GRAD-Net using the Oxford5k dataset. Fig. \ref{fig:fl} plots the t-SNE visualizations of the Oxford5k images using the original image descriptors and the learned features of GRAD-Net respectively. The nodes are colored by the ground truth labels. Fig. \ref{fig:fl} shows that GRAD-Net groups the instances from the same category with similar features.

\begin{figure}[h!]
\centering
\includegraphics[width=0.45\textwidth]{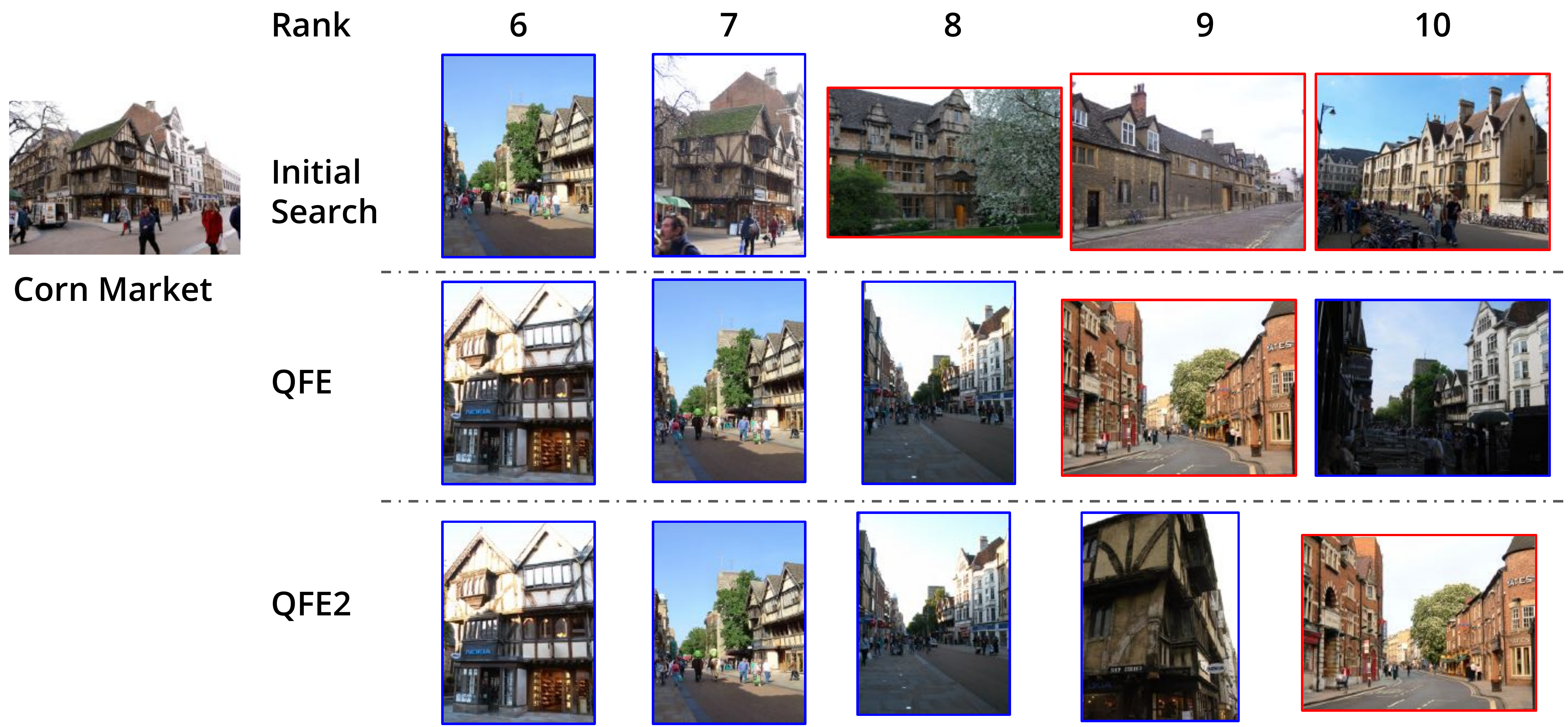}
\caption{{\bf Query Induction Example.} The first row represents the initial search result in the original feature space. The second row represents the first QFE search result, and the third row is the second QFE search result, which implies QFE can improve the retrieval results for unseen queries.}
\label{fig:qe}
\end{figure}

\begin{figure*}[h!]
\centering
\includegraphics[width=1\textwidth]{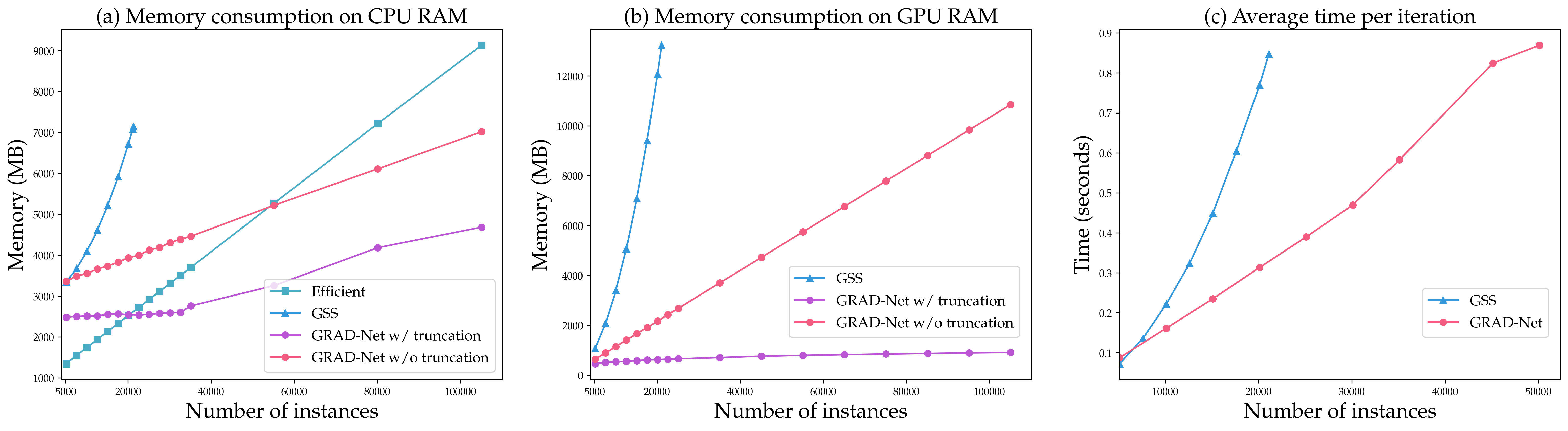}
\caption{{\bf Memory and time consumption comparison.} (a) CPU RAM memory consumption against the number of dataset instances; (b) GPU RAM memory consumption against the number of dataset instances. Efficient is omitted since the method cannot be implemented in GPU. (c) Time consumption against the number of instances.} 
\label{fig:complexity}
\end{figure*}

\subsection{Inductive Leaning} \label{sec:il}
To test the generalizability of GRAD-Net for unseen queries, we manually put aside 11 queries from the 55 queries of the Oxford5k dataset and left them out in the training. We then performed retrieval of these 11 queries using the QFE method described in Section \ref{sec:ilui}. The mAP score in the Table \ref{tb:il} shows the effectiveness of the QFE method. We also observe a slight improvement in the performance when applying QFE twice. As shown in Fig. \ref{fig:qe}, the QFE method can project approximated features of the unseen query to the learned feature space, which leads to more accurate retrieval performance.

\begin{table}[h!]
\renewcommand{\arraystretch}{1.3}
\caption{Performance comparison on unseen queries}
\label{tb:il}
\centering
\begin{tabular}{|c|c|c|c|c|}
\hline
Method & Model & mAP  \\
\hline
Query Expansion & Efficient \cite{yang2019efficient} & 89.6\\
\hline
QFE & \multirow{2}{*}{GRAD-Net} & 96.01\\
QFE $\times$ 2 & & 96.05\\
\hline
\end{tabular}
\end{table}

\subsection{Complexity Analysis} \label{sec:ca}
The first step in the pipeline of GRAD-Net is to build a mutual k-NN graph, which requires the pairwise similarity search among instances.
This step has a computational complexity of $O(N^2)$, where $N$ is the number of instances. However, since we only need to build the graph once and it is considerably accelerated by the faiss toolbox \cite{johnson2019billion}, we argue that the computational complexity of GRAD-Net is $O(N)$ that is mainly determined by the training process of the neural network. 

We conducted a series of experiments to evaluate the memory and time consumption of GRAD-Net and several other methods. All experiments were conducted using a workstation with a 12-core Intel(R) Xeon(R) CPU W-2133 @ 3.60GHz and one NVIDIA RTX5000 GPU with 16GB GPU memory. We started with the Oxford5k dataset, gradually added distractors to the original dataset, and recorded the time and memory consumption of training or the diffusion iterations. Empirical results verify our argument as follows.

\noindent \textbf{Memory consumption analysis.} Fig. \ref{fig:complexity}(a-b) show the memory consumption on CPU and GPU respectively. As the number of instances increases, GRAD-Net can easily scale up to 10k samples even without using truncation, thanks to the mini-batch training on graphs. After integrating the truncation technique, the memory consumption is notably reduced by a large margin. This validates the advantage of GRAD-Net regarding memory consumption.

In comparison, the memory consumption of Efficient Diffusion \cite{yang2019efficient}, a conventional diffusion method, exceeds GRAD-Net as the number of instances reaches 20k regardless of that it employs a truncated search technique.
We also compare with the memory consumption of GSS \cite{liu2019guided}, a neural network based method, increases rapidly as the number of instances gets larger, and encounters \textit{out of memory (OOM)} error when the number of instances exceeds 21k. 

\noindent \textbf{Time consumption analysis.} We analyzed the time consumption of GRAD-Net as the number of instances increases. Fig. \ref{fig:complexity}(c) plots the average time per forward computing and gradient back-propagation against the numbers of instances. The time consumption results validate that GRAD-Net has linear complexity, $O(N)$. We also compare GRAD-Net with the neural network based method, GSS (blue line in Fig. \ref{fig:complexity}(c)), and argue that GRAD-Net has faster iteration time.

\section{Conclusion}
In this paper, we propose GRAD-Net, a novel deep learning based diffusion method to address the challenge of unsupervised image retrieval. In contrast to conventional diffusion methods, GRAD-Net enables effective representation learning for images while preserves both the local and global geometric properties of the image manifold. Moreover, GRAD-Net is trained in an unsupervised end-to-end fashion and can easily scale up to handle large-scale datasets. Extensive empirical results on multiple benchmarks demonstrate the efficacy of GRAD-Net. Besides its outstanding performance, GRAD-Net also shows its generalizability to unseen queries in an inductive learning manner. However, for highly dynamic image retrieval systems where instances are constantly modified in the database, we envision that GRAD-Net is capable of performing the retrieval tasks with regular re-training. The trigger to re-train the network can be determined by the mean average precision over a separate validation set. We will explore these directions in our future works. 

\ifCLASSOPTIONcaptionsoff
  \newpage
\fi

\printbibliography
\begin{IEEEbiography}[{\includegraphics[width=1in,height=1.25in,clip,keepaspectratio]{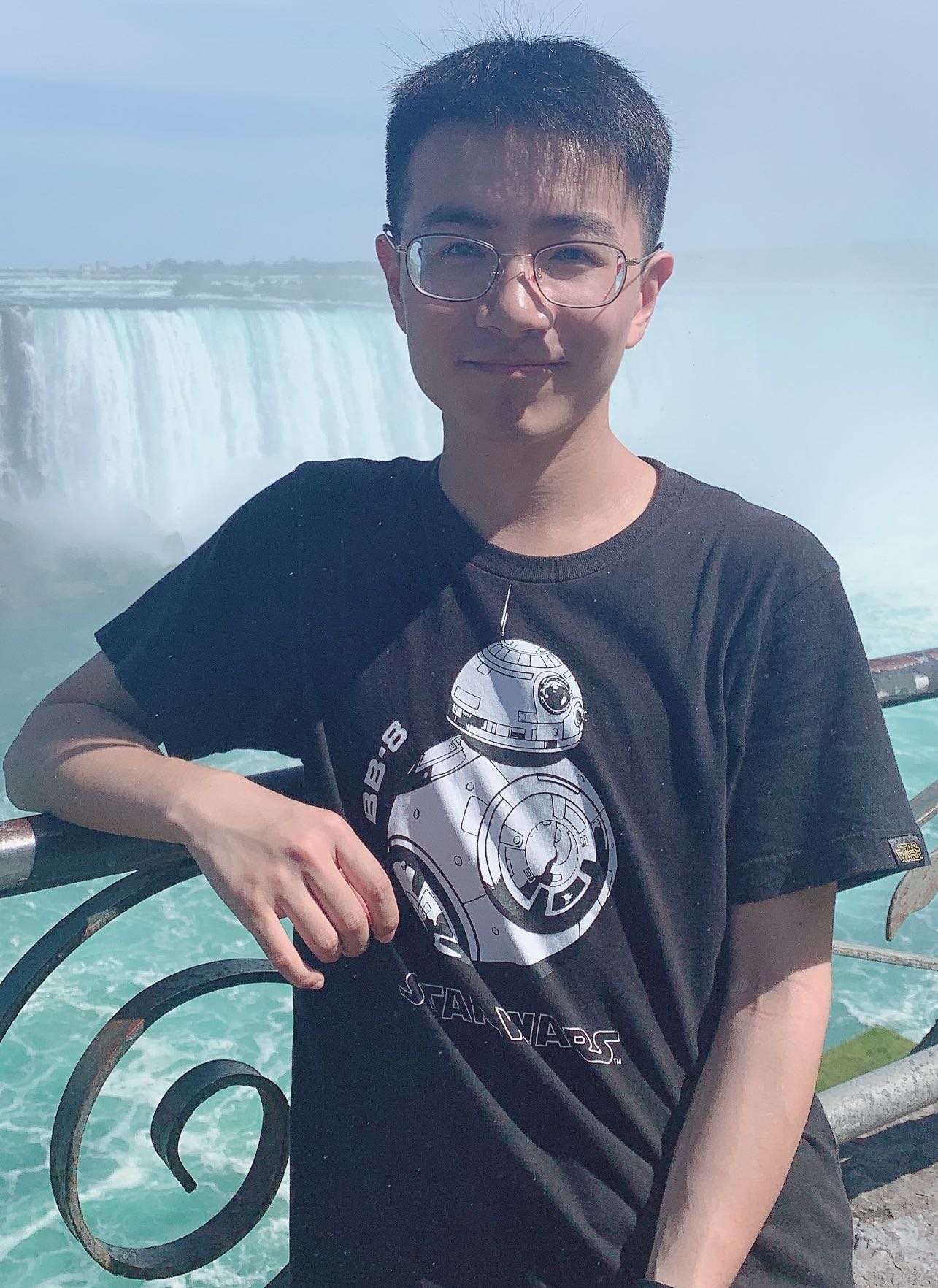}}]{Zhiyong Dou} received his B.S. in Electric and Electronic Engineering from Huazhong University of Science and Technology, Wuhan, China in 2017. He is currently a Ph.D candidate at the School of Electronic Information and Communications in Huazhong University of Science and Technology. He is also a visiting student at the University of Toronto beginning in 2019. His research interests include computer vision, computational biology and machine learning.
\end{IEEEbiography}

\begin{IEEEbiography}[{\includegraphics[width=1in,height=1.25in,clip,keepaspectratio]{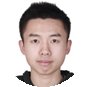}}]{Haotian Cui} received his B.S. in Biomedical Engineering from the
Tsinghua University, China in 2015. He is currently pursuing his Ph.D. degree in Computer Science at the University of Toronto. His research interests include computer vision, computational biology and machine learning.
\end{IEEEbiography}
\begin{IEEEbiography}[{\includegraphics[width=1in,height=1.25in,clip,keepaspectratio]{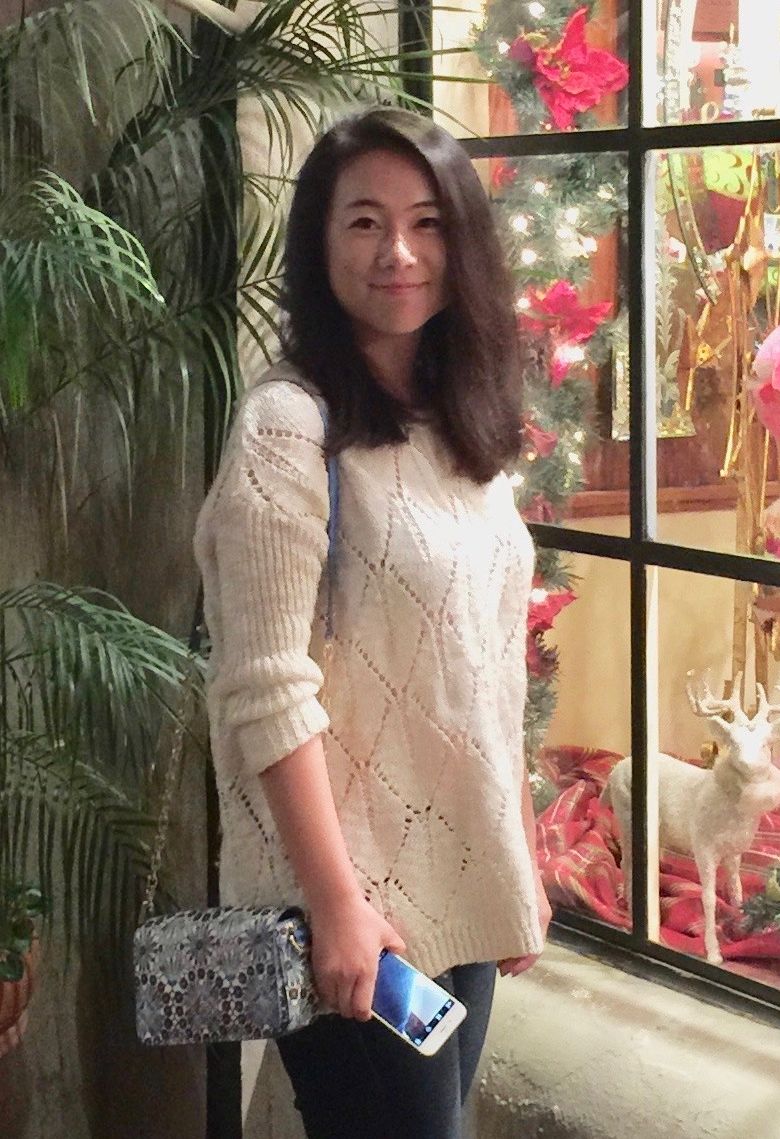}}]{Lin Zhang} 
received her HB.A. in Statistics and B.A. in Economics from University of California, Berkeley in 2012, and received her M.A. in Applied Statistics from University of California, Santa Barbara in 2015. She is currently a Ph.D. candidate at the Department of Statistical Sciences at the University of Toronto. 
\end{IEEEbiography}

\begin{IEEEbiography}[{\includegraphics[width=1in,height=1.25in,clip,keepaspectratio]{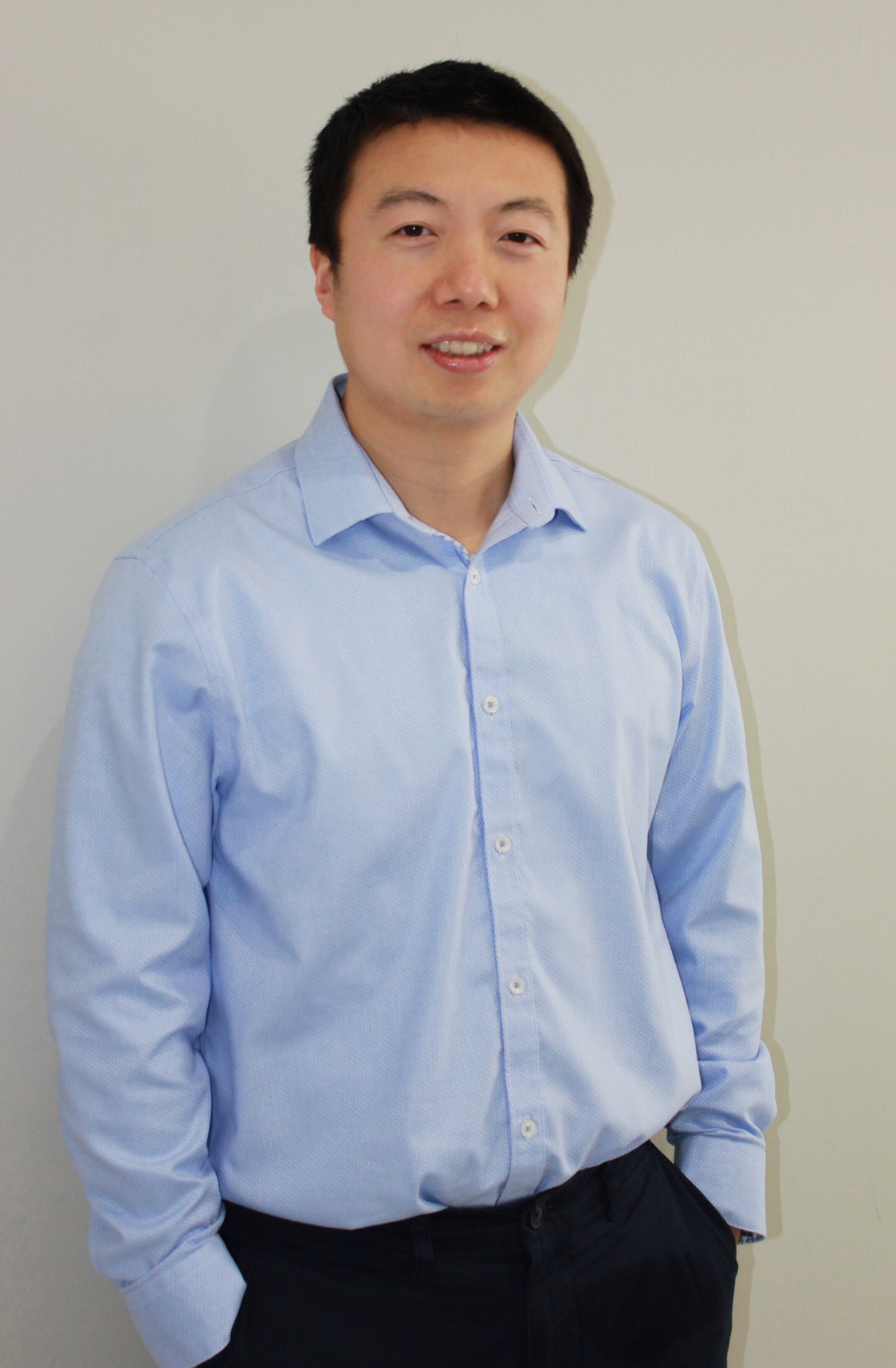}}]{Bo Wang} is the lead AI scientist for the Peter Munk Cardiac Centre at the University Health Network (UHN) in Toronto. He is also an Assistant Professor at the Department of Medical Biophysics at the University of Toronto and a CIFAR AI Chair of the Vector Institute. Dr. Wang obtained his Ph.D. from the Department of Computer Science at Stanford University in 2017 and his M.Sc. in Computer Science from the University of Toronto in 2012. His current research interests include computer vision, computational biology and machine learning. Dr. Wang is a member of IEEE.
\end{IEEEbiography}

\end{document}